# Edge Computing for Real-Time Near-Crash Detection for Smart Transportation Applications

Ruimin Ke, *Member, IEEE*, Zhiyong Cui, *Member, IEEE,* Yanlong Chen, Meixin Zhu, Hao (Frank) Yang, *Student Member, IEEE,* and Yinhai Wang, *Senior Member, IEEE*

*Abstract*—Traffic near-crash events serve as critical data sources for various smart transportation applications, such as being surrogate safety measures for traffic safety research and corner case data for automated vehicle testing. However, there are several key challenges for near-crash detection. First, extracting near-crashes from original data sources requires significant computing, communication, and storage resources. Also, existing methods lack efficiency and transferability, which bottlenecks prospective large-scale applications. To this end, this paper leverages the power of edge computing to address these challenges by processing the video streams from existing dashcams onboard in a real-time manner. We design a multi-thread system architecture that operates on edge devices and model the bounding boxes generated by object detection and tracking in linear complexity. The method is insensitive to camera parameters and backward compatible with different vehicles. The edge computing system has been evaluated with recorded videos and real-world tests on two cars and four buses for over ten thousand hours. It filters out irrelevant videos in real-time thereby saving labor cost, processing time, network bandwidth, and data storage. It collects not only event videos but also other valuable data such as road user type, event location, time to collision, vehicle trajectory, vehicle speed, brake switch, and throttle. The experiments demonstrate the promising performance of the system regarding efficiency, accuracy, reliability, and transferability. It is among the first efforts in applying edge computing for real-time traffic video analytics and is expected to benefit multiple sub-fields in smart transportation research and applications.

*Index Terms*—Edge computing, intelligent vehicle, near-crash detection, real-time video analytics, smart transportation

## I. Introduction

NEAR-crash, or near-miss, is an incident that would have resulted in a loss such as property damage or injury. Herbert Heinrich proposed the relationship among major injury, minor injury, and no injury incidents (1 major injury incident to 29 minor injury incidents to 300 no injury incidents) [1]. Under the context of road transportation, a near-crash is a traffic conflict between road users that has potential to develop into a collision. Moreover, the linear relationship found by Heinrich still holds, though the ratio numbers could be different [2]–[7]. Near-crash has two major properties that make it valuable for a variety of research and engineering topics: (1) It reflects the underlying causes of the incidents while results in no or minor losses; (2) It is in a much larger number than the real accidents.

Near-crash data is irreplaceable in smart transportation applications. In traffic safety research, near-crash data is the surrogate safety data for studying and assessing the safety performance of certain locations or scenarios [8]–[11]. This is because a certain amount of data is required to feed either traditional statistical analytical methods or emerging machine learning models. For example, to understand the safety-related designs at a roadway intersection, the collision data might be far from sufficient to support models to reach any statistically significant conclusions. Near-crash data fills this hole with the aforementioned two properties.

With the emergence of concepts and technologies in the intelligent vehicle (IV) and autonomous vehicle (AV), near-crash becomes an even more valuable data source for not only traditional traffic safety research but also IV and AV safety. The latest AVs have been demonstrated to be able to handle most situations they may encounter. However, the lack of corner cases for training and testing is a major bottleneck that is slowing down the pace to achieve the goal of Level-5 (L5) fully autonomous driving [12], [13]. Corner cases belong to subsets of near-crashes; they rarely occur, such as pedestrians walking across a highway, but can cause severe losses. Leading research in the AV field is focused on speeding up the generation of corner cases in simulation by leveraging the historical crash or near-crash data for model calibration and training [14].

There are generally two sources of sensors for near-crash data collection, roadside sensors and vehicle onboard sensors. Surveillance cameras are the commonly used roadside sources for near-crash data collection given their properties of being widely deployed and information-rich. The City of Bellevue has been leading a collaborative effort with Microsoft Research and the University of Washington on large-scale near-crash data collection using city-wide surveillance cameras towards achieving the mission of Vision Zero [15]. Vehicle onboard sensors help cover the areas where no roadside sensors are installed. In a recent pilot project, near-crash data were collected using dashcams for the evaluation of Mobileye Shield+ system and the its potential to reduce transit bus collisions with pedestrians [16].

Three key challenges remain for near-crash detection and

This work was supported in part by the Safety Research and Demonstration (SRD) grant from the Federal Transit Administration (FTA) and in part by the Pacific Northwest Transportation Consortium (PacTrans).

Ruimin Ke is with the Department of Civil Engineering, University of Texas at El Paso, El Paso, TX, USA. Zhiyong Cui, Meixin Zhu, Hao (Frank) Yang, and Yinhai Wang are with the Smart Transportation Applications and Research (STAR) Lab, Department of Civil and Environmental Engineering, University of Washington, Seattle, WA, USA. Yanlong Chen is with the Department of Mechanical Engineering, University of Tokyo, Bunkyo City, Tokyo, Japan. *Corresponding author: Yinhai Wang (email: yinhai@uw.edu)*

data collection. Firstly, near-crashes are still rare events which require intensive computing, transmission bandwidth, and data storage services on the original large data sources. Secondly, existing methods documented in papers and reports rely on manual checking or post-analysis on the original data sources, which are inefficient [6], [7], [9], [17]. Thirdly, while the state-of-the-practice commercial collision avoidance systems (e.g., the aforementioned Shield+ system) can serve for the purpose of near-crash data collection, their purchase and maintenance costs are very high, and each product lacks transferability and scalability due to the design for certain type of vehicles or roadside units.

These challenges all limit the deployment scale and speed of near-crash data collection. An emerging term, edge computing, appears to be a natural solution. Edge computing targets processing data closer to where the data is generated, thus it can offload the computation and storage burdens on the cloud, save transmission volume and bandwidth, and better protect privacy [18]. In the potential large-scale deployment of near-crash detection systems (e.g., thousands of vehicles), it is not feasible to either store all the raw videos locally due to the storage limit or transmit all raw videos in real-time due to the bandwidth limit. An ideal system for near-crash data collection is low-cost by using existing dashcams, transferrable to different vehicles, and running in a real-time manner.

Under the context of edge computing, real-time video analytics is regarded as the Killer App [19], given the restricted computing resources on edge devices and the property of video data being in large volume as enormous 3D matrices. To this end, this paper introduces a light-weight edge computing system for real-time near-crash detection and data transmission with normal dashcams and network bandwidth. The system is a low-cost and standalone system that is backward-compatible with existing vehicles. It is developed based on the Nvidia Jetson TX2 Internet-of-Things (IoT) platform.

Algorithm-wise, there are three major innovations. Firstly, the proposed near-crash detection algorithm models the output bounding boxes from state-of-the-art deep-learning object detectors for Time-To-Collision (TTC) and horizontal motion estimation with a linear complexity, which is both accurate and efficient. Secondly, with mathematical proof, the TTC value, which is the most widely used indicator for near-crash detection, is not sensitive to camera parameters using the proposed algorithm, thus ensures great transferability to any dashcam on existing vehicles. Thirdly, new rules are defined for near-crash identification using the TTC and horizontal motion estimations.

System-wise, this study proposes a multi-thread system that handles video reading, near-crash detection, data transmission, and data fusion for associated near-crash events. This system architecture is able to handle all the functional modules in real-time on NVIDIA Jetson. The system processes all the videos on the network edge nodes, filter out most of the irrelevant videos, and transmitting only the relevant videos, Controller Area Network (CAN) data (e.g., brake switch, vehicle speed, throttle, decelerations), GPS data that are associated with the near-crashes to the cloud server.

The edge computing system for near-crash detection has been comprehensively tested using online videos, real-world tests on two cars and four transit buses for over ten thousand hours. Evaluation and analysis on the results demonstrate promising performance of the proposed system regarding accuracy, efficiency, and the richness of output near-crash data. A short demo video is published online at https://www.youtube.com/watch?v=8qu-cNqfWkg.

## II. UNDERSTANDING RELATIVE MOTION PATTERNS IN CAMERA FOR NEAR-CRASHES

Relative motions between the ego-vehicle and other road users are important cues for near-crash detection using a single camera [7], [20]. Relative motion patterns as well as the relationship between a pattern in the camera view and its corresponding pattern in the real world must be understood (see Figure 1). The relative motion patterns between two road users vary from case to case. Roadway geometry, road user's behavior, relative position, traffic scenario, etc. are all factors that may affect the relative motion patterns. For example, from the ego-vehicle's perspective, its motion relative to a vehicle it is overtaking in the neighbor lane and that to another vehicle it is following in the same lane are different.

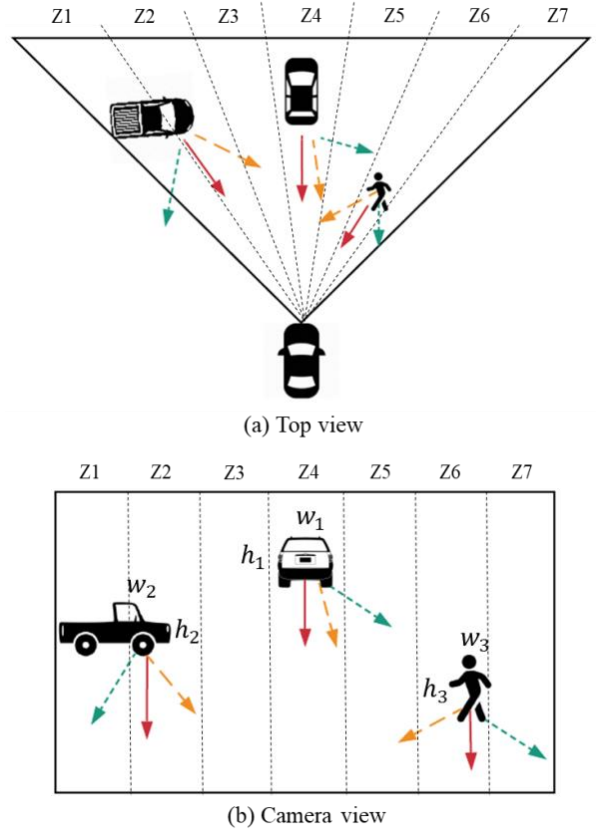

Fig. 1 The corresponding relative motions, relative locations, and lines of sights between the ego-vehicle and three other target road users. In the case of target's size increasing in the camera view, there are still three types of relative motions between the ego-vehicle and the target road user — solid red arrows: potential crashes; dotted yellow arrows: warnings; dotted green arrows: safety.

Relative motion that has the potential to develop into a crash/near-crash is characterized from the ego-vehicle's perspective as the target road user moving towards it. This kind of relative motion is shown as a motion vector of the target road user moving vertically towards the bottom side of the camera view. Examples are shown as solid red arrows in Figure 1. In the real-world top view, the three solid red arrows represent the relative motions between the ego-vehicle and each of three road users (a pick-up truck, a car, and a pedestrian). Each of the three camera sight lines aligns with a relative motion vector (Z2, Z4, and Z7). In the camera view, the lines of sight are shown as vertical bands. The relative motion vectors for near-crashes in the top view correspond to vectors moving towards the bottom in the camera view aligning with Z2, Z4, and Z7.

Two road users have a relative motion at any time. In addition to the near-crash cases defined above, other patterns may occur. First, a target road user may move towards the ego-vehicle, move away from the ego-vehicle, or stay at the same distance to the ego-vehicle. These can be identified as object image size changes in the camera. This property will be utilized later in our approach. Image size decreasing or no size change would not indicate a potential crash or near-crash. For size increasing, there are three cases. The first cases are the potential crashes, shown as the solid red arrows in Figure 1. The second are the warning cases, shown as the dotted orange arrows, in which the relative motion is towards the center line of sight of the camera (the pick-up truck and the pedestrian), or the relative motion is slightly different from the solid red arrow while the target road user is at the center line of sight (the car). The warning cases could develop into crashes if there are slight changes in the speeds or headings of either the target or the ego-vehicle. The third case is the safety case that relative motion is moving away from the center line of sight, shown as the dotted green arrows in Figure 1.

## III. EDGE COMPUTING SYSTEM ARCHITECTURE

The overall system architecture on the edge computing platform is shown in Figure 2. The two major functions of the system are near-crash detection and data collection. Given the real-time operation requirement for both functions, the design should be simple enough to support to be highly efficient and sophisticated enough to use the Nvidia Jetson's computational power for high accuracy and reliability. The near-crash detection method also should be insensitive to camera parameters to accommodate large-scale deployments.

The system is implemented in a multi-thread manner. Four different threads are operating simultaneously: main thread, data transmission thread, video frame reading thread, and CAN receiving thread (CAN = Controller Area Network). The proposed near-crash detection method is implemented in the main thread. When near-crash events are detected, a trigger will be sent to the data transmission thread, and it will record video frames from a queue (a global variable) and other data that are associated with the near-crash event. The third thread for video frame reading keeps the latest video frame captured from the camera in another queue and will dump previous frames when the capturing speed is faster than the main thread's frame processing speed. The CAN receiving thread provides additional information for each near-crash event with the ego-vehicle's speed, brake, acceleration, and so forth.

The proposed architecture ensures that the system delay is low. The frame reading thread ensures that the main thread reads the latest frame captured by the camera by not accumulating frames. The data transmission thread is designed as an individual thread to handle data transmission so that the main thread operation is not affected by the network bandwidth. The CAN receiving thread is for additional information collection, and the purpose for separating it as another individual thread is the consideration of system function extension. The proposed system can communicate with other systems via this thread while not affecting the performance of itself.

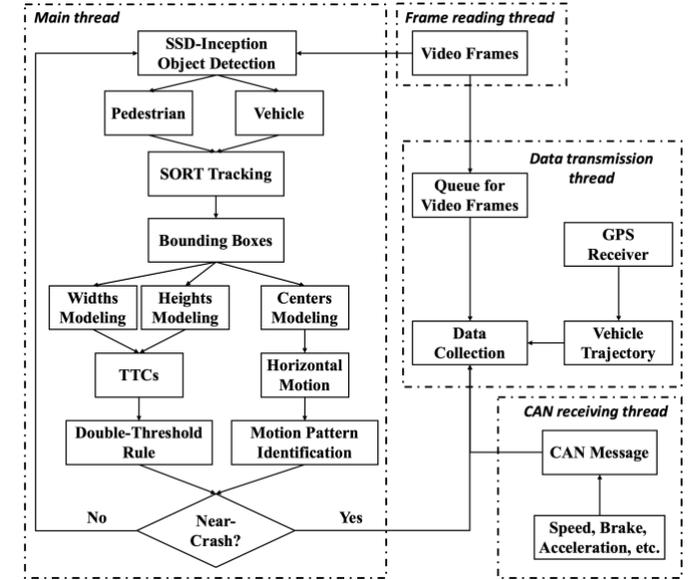

Fig. 2 The system architecture on the edge computing device

## IV. REAL-TIME CAMERA-PARAMETER-FREE NEAR-CRASH DETECTION ALGORITHM

### A. Deep-learning-based road user detection and tracking

The main thread starts with applying a deep-learning-based object detector to every video frame. Deep-learning-based object detection can simultaneously localize and classify objects with high accuracy [21], [22]. However, one disadvantage of deep-learning-based inference is its high computational cost, which prevents it from being deployed for certain applications. As one of the most powerful IoT devices in the past few years, Nvidia Jetson TX2 is capable of running some deep object detectors in real-time and running the inference with TensorRT-optimized inference neural networks.

For traditional IoT devices, Tiny YOLO (You Only Look Once) and SSD (Single Shot Multibox Detector)-Mobilenet are two of the most popular deep-learning-based detectors given their high inference efficiency. However, we chose a more complicated detector, SSD-Inception, a real-time detector on Jetson TX2 with nearly 30 frames-per-second (FPS) detection speed and better accuracy. The system keeps the bounding boxes of the detected pedestrians and vehicles for further processing.

The object detection creates bounding boxes and identifies the types of road users in each video frame. To associate the information from each frame and find each road user's movement, a standard step following object detection is object tracking. SORT (Simple Online Real-time Tracking) tracking [23] is a recent benchmark for object tracking with online and real-time performance. It achieves good tracking accuracy without the need for any complicated features but solely the bounding box information. It also can eliminate some false-positives and false-negatives that are generated in the detection phase. Some studies demonstrated it to be a suitable tracking method for intelligent transportation applications [24], [25].

*B. Modeling bounding boxes in linear complexity for camera-parameter-free TTC estimation*

An object appears larger in the camera view as it is approaching the camera, and smaller as distance to the camera increases. Researchers at Mobileye published a paper as early as 2004 to show that it was possible to determine TTC using size changes [26]. In this study, the proposed approach for TTC estimation mainly considers: (1) leveraging the power of recent achievements in deep learning, (2) making the computation as efficient as possible to support real-time processing on Jetson, and (3) transferability to any dashboard camera without knowing the camera's intrinsic parameters.

Use of deep learning was discussed in the last sub-section. SSD+SORT detects and tracks road users with high accuracy. However, the next step, which is the near-crash identification, must be simple and effective. Otherwise, the real-time requirement would not be satisfied by the IoT device.

Object detection and tracking provide the locations, categories, and sizes of objects in the bounding boxes information. However, bounding boxes are just approximate sizes of the objects and are not used for accurate determination of object size. Particularly, given two consecutive frames, the size change of an object is subtle; and in many cases, this change is not recognizable due to noise in the bounding box generation. In our initial experiment, we also found that the size of an object in the previous frame may be even larger than that in the next frame.

Another reason for inaccurate size change detection in neighboring frames is that the time is too short in between two consecutive frames. Given a video with a frame rate of 24 FPS, the next frame is captured in less than 0.05 seconds. Thus, for size change detection we use more frames to compensate for the noise in each frame and increase the time interval for the detection. Linear regression is used for bounding boxes' heights or widths over a group of consecutive frames. We found that 10 to 15 frames are enough to compensate for noise and the time associated with 10 to 15 frames is still small enough (about 0.5 second) to assume that the road user's motion is consistent.

Therefore, the input to the linear regression is a list of heights or widths extracted from the bounding boxes, and the slope outputted by the regression will be the size change rate.

Let us denote the size change rate as $r_t$, and the size of the road user in the video frame as $s_t$ at time $t$. At the same time, in the real world, the longitudinal distance between the target road user and the ego-vehicle is $D_t$, the relative longitudinal speed is $V_t$, the target road user's size is $S_t$, and the camera focal length is $f$. Based on the pinhole camera model, there is

$$\frac{s_t}{f} = \frac{S_t}{D_t} \qquad (1)$$

Relative speed is the first derivative of relative distance, and that size change rate is the first derivative of the object size over time

$$V_t = \frac{dD_t}{dt}, \qquad r_t = \frac{ds_t}{dt} \qquad (2)$$

Since the real-world target road user's size does not change over time, there is the following equation

$$0 = \frac{dS_t}{dt} = \frac{d\left(\frac{D_t s_t}{f}\right)}{dt} \qquad (3)$$

And since the focal length does not change over time, we have

$$0 = \frac{d(D_t s_t)}{dt} = \frac{dD_t}{dt} \cdot s_t + \frac{ds_t}{dt} \cdot D_t = V_t s_t + r_t D_t \qquad (4)$$

Thus,

$$TTC = -\frac{D_t}{V_t} = \frac{s_t}{r_t} \qquad (5)$$

According to Eq. (5), TTC can be calculated as the size of the bounding box at time $t$ divided by the size change rate at time $t$. It is not related to the focal length or other intrinsic camera parameters. The TTC value can be either positive or negative, where being positive means the target is approaching the ego-vehicle, and being negative means it is moving away from the ego-vehicle.

*C. Height or width?*

There are two options for the size of the road user in the camera view, height and width. We argue that height is a better indicator than width. From the ego-vehicle's perspective, it may observe a target vehicle's rear view, front view, side view, or a combination of them, depending on the angle between the two vehicles. That is to say, the bounding box's width change may be caused by either the relative distance change or the view angle change. For example, when the ego-vehicle is overtaking the target vehicle, or the target vehicle is making a turn, the view angle changes and will lead to the bounding box's width change.

However, the bounding box's height of the target vehicle is not influenced by the view angle; it is solely determined by the relative distance between the two vehicles. Similarly, a pedestrian walking or standing on the street may have different bounding box widths due to not only the relative distance to the ego-vehicle but also the pose of the pedestrian; but the height of a pedestrian is relatively constant.

Despite the challenge of using width to determine an accurate TTC, it still provides valuable information. Since we are using only less than one second of frames for the calculation, the view change does not contribute as much as the distance change, so width still roughly shows the longitudinal movement of the road user. This is very important in some cases. For instance, a vehicle moving in the opposite direction of the ego-vehicle is truncated by the video frame boundary. In this case, the height of the vehicle increases while the width decreases. This is not a near-crash case at all, but the TTC can be very small and falsely indicate a near-crash by only looking at the height change.

We propose a double-threshold rule: if the TTC threshold for determining a near-crash is $\delta$, we will set this $\delta$ as the TTC threshold associated with the height regression. At the same time, we have another TTC threshold $\varphi$ associated with the width regression. The second threshold $\varphi$ is to ensure that the width and height changes are in the same direction. The rule is represented as

$$0 < \frac{h}{r_h} < \delta, \qquad 0 < \frac{w}{r_w} < \varphi, \qquad \delta < \varphi \qquad (6)$$

where $r_h$ and $r_w$ are the change rates for height $h$ and width $w$. It is a necessary condition for a near-crash.

*D. Modeling bounding box centers for horizontal motion pattern identification*

As shown in Figure 1, there are three scenarios for the case that a road user approaches the ego-vehicle; they correspond to potential crashes, warnings, and safe scenarios. Besides TTC, these scenarios can be differentiated with the relative horizontal motion between the ego-vehicle and the target. This needs to be calculated with computationally fast methods as well. We propose to apply another linear regression using a list of bounding box's centers of the target road user. The regression result would be able to indicate the moving direction of the road user in the camera view.

In general, when the target's location is closer to the bottom and closer to the center line of sight, the risk of a collision is higher, so the threshold for the moving direction $\omega$ is looser. We propose a rule to show this judgment as

$$\alpha < \omega \cdot (C_x - C_{los}) \cdot (B_y - B) < \beta \qquad (7)$$

where $C_x$ is the center's x coordinate, $C_{los}$ is the center line of sight, $B_y$ is the bottom side of the bounding box, and $B$ is the bottom of the video frame. Since cameras have different resolutions, $(C_x - C_{los})$ is normalized to [-1, 1] and $(B_y - B)$ is normalized to [0, 1]. The two thresholds are $\alpha$ and $\beta$; $\alpha$ should be set to negative to capture the potential warning scenarios (the orange dotted arrows in Figure 1). And $\beta$ should be just slightly larger than zero to capture the potential crashes (the solid red arrows in Figure 1) and filter out most of the safe scenarios (the green dotted arrows in Figure 1). Eq. (6) and Eq. (7) together identify near-crash events.

V. EXPERIMENTAL RESULTS AND ANALYSIS

*A. Experiment Design*

Local experiments with locally stored videos at Jetson and real-world experiments with onboard real-time video feeds were selected as two groups for testing the system. Local video resources covered a lot of historical near-crash scenarios as well as other corner cases. It was a better source to evaluate the near-crash detection method we proposed in this paper. Real-time video stream data was captured by the system on cars and buses. Over 1000 hours of tests have been conducted so far. Local video data were also collected from online sources (e.g., YouTube) and dashboard cameras. Real-world tests have been conducted on two Honda cars and four Pierce Transit buses for over six months in the year 2020 and 2021. Figure 3 shows the system and testing buses for the real-world test. From top to bottom: the systems ready to be installed (before installation), three of the testing buses at Pierce Transit, the radio box behind bus driver's seat where the system works, and the system being tested in the radio box.

*B. Hardware component*

The system consists of an Nvidia Jetson TX2, a dashcam (can be USB camera or IP camera), a GPS receiver, an in-vehicle power inverter, a PEAK CAN adapter for CAN bus communication, an external circuit based on Arduino board for auto bootup, a shell for the Jetson device, an ethernet cable, two power cables, an internet switch, mounting materials, and a cloud server. The Nvidia Jetson device is the key processing unit of the system, running the near-crash detection, video streaming, data transmission, data fusion threads and algorithms. The Jetson was powered by in-vehicle (either car or bus) 12V DC power through the power inverter. The Arduino circuit is connected to the Jetson, and when the vehicle's power is on, it will auto boot up the system.

*C. Parameter Settings*

Several key parameters needed to be set properly: SSD detector confidence threshold, the number of frames for size regression, the number of frames for center regression, TTC threshold $\delta$, TTC threshold $\varphi$, horizontal motion threshold $\alpha$, horizontal motion threshold $\beta$, and Jetson power mode. Given that the SSD detector tended to have fewer false-positives than false-negatives [24], some false-positives can be filtered out at the tracking step, and more false-positives (if any) will be filtered out by the near-crash detection algorithm, we set the detection confidence threshold to be 0.3 – 0.5.

For the number of frames for size regression, we suggested setting them to be around 10 to 15 frames. This range was large enough to compensate for the bounding box noises and small enough to assume the target's motion is consistent. The number of frames for center regression can be a little larger to capture the horizontal motion better, and the suggested number was in the range of 15 to 20. For $\delta$ and $\varphi$, as defined by many previous studies, the TTC threshold for a near-crash was around 2 to 3 seconds, which was our suggested value for $\delta$. And we found that setting $\varphi$ to about 2 to 2.5 times $\delta$ worked well. We suggested setting $\alpha$ to the range of [-1, -0.5] and $\beta$ to [0.02, 0.1]. Jetson power mode was recommended to be set as Max-N to fully utilize its computational power, though our system still operated in real-time (but lower FPS) with Max-Q mode.

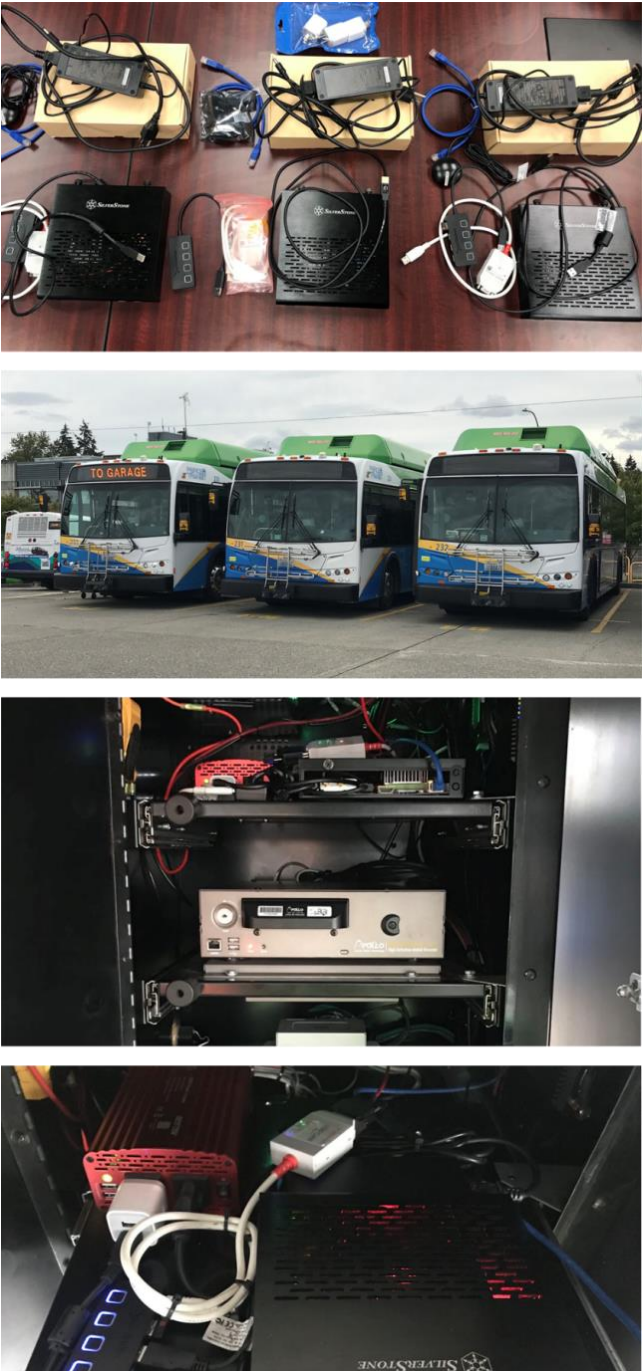

Fig. 3 The system prototypes, buses for the real-world testing, and the bus radio box where the system works.

### D. Evaluation of Near-Crash Detection

Essentially, near-crash is a type of traffic anomaly. To evaluate the proposed method's accuracy, we used the evaluation process of the Traffic Anomaly Detection task (Track 4) of the 2020 AI City Challenge as the reference [27]. First, the task dataset has 100 video clips with some anomalies. It is unknown exactly how many anomalies are in the test dataset, but the number is between 0 and 100, as mentioned in the introduction to Track 4. Likewise, we made a local test dataset with 5000 video clips with 500 near-crash events. As aforementioned, the test videos were from online resources, dashboard cameras on private cars and transit buses. This dataset is not being published due to potential privacy and copyright issues. There is a plan to create such a video dataset for near-crash detection in the future.

We manually labeled all the near-crash events with their occurrence videos and times. As in AI City Challenge Track 4, we defined a true-positive (TP) as a predicted near-crash within 10 seconds of the true near-crash. A false-positive (FP) is a predicted near-crash that is not a TP for a near-crash. A false-negative (FN) was a true near-crash that was not predicted. We used the F1 score to evaluate accuracy. F1 score was the harmonic mean of the precision and recall, where the best value = 1 and the worst value = 0.

$$F1 = 2 \cdot \frac{precision \cdot recall}{precision + recall} = \frac{2TP}{2TP + FP + FN} \quad (8)$$

Sample near-crash detection results are shown in Figure 4. The top three rows were three vehicle-vehicle near-crashes, and the bottom two rows were two of the vehicle-pedestrian near-crashes. The bounding boxes turned red to indicate a predicted near-crash, while other detected road users had green bounding boxes. A few more sample detection results can be found in the video published at https://www.youtube.com/watch?v=8qu-cNqfWkg.

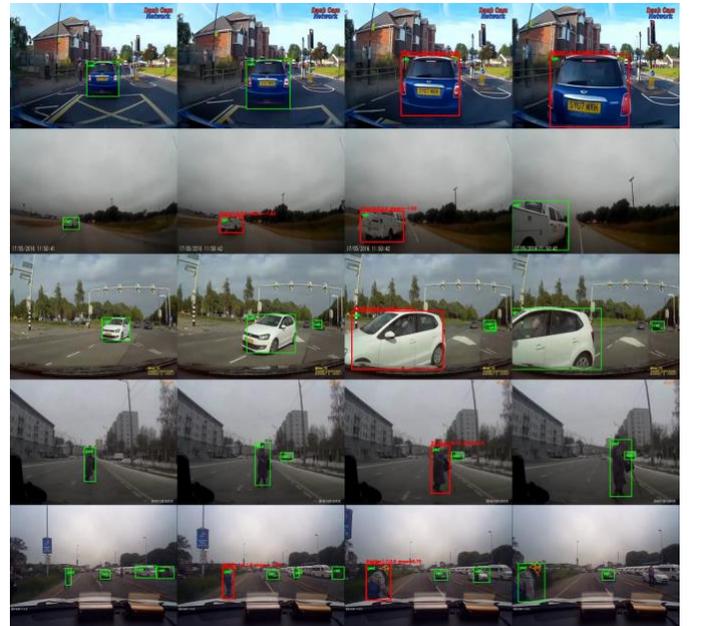

Fig. 4 Sample near-crash detection results, where red bounding boxes indicate the potential conflict with the road user. Each row is a four-frame sequence of one near-crash event.

As summarized in Table I, our system correctly predicted 496 out of the 500 labeled near-crashes and missed just 4. It generated 8 FPs in the 5000 video clips. Based on Eq. (8), the final F1 score was 0.988, and the average processing speed with Max-N mode was about 18 frames-per-second (FPS). The performance was promising, considering that we intentionally included a variety of near-crash scenarios and some very challenging cases in the dataset. There were adverse weather conditions (e.g., foggy, rainy, snowy), nighttime situations, traffic congestion, urban/rural traffic scenes, and so on. It is worth mentioning that the 5000 video clips are from a lot of different cameras and the proposed system knew nothing about the camera parameters of any of these cameras. This result benefited from the near-crash detection method. It again highlighted the possibility for low-cost and highly efficient large-scale application of the edge computing system to partially fulfill the purposes of safety data generation, IV corner case collection, and collision avoidance.

We carefully examined the FN and FP cases and summarized the causes. One of the four FNs that the system missed was a vehicle-pedestrian near-crash at night on a rural freeway with no streetlight. The pedestrian violated traffic rules by crossing the freeway, and the driver did not see him until almost running into him. The pedestrian was entirely in the dark so that the object detector missed him. Though there were more FPs than FNs, we considered only 8 FPs out of 5000 video clips acceptable and encouraging given the tradeoff in the efficiency of the system. While the proposed near-crash detection method can compensate for bounding box size noise in most cases, it was not perfect. In the fourth case (the fourth row) of Figure 4, right before the correct detection of this vehicle-pedestrian near-crash, there was a vehicle-vehicle FP caused by a significant error in vehicle size detection. It was included in our YouTube demo video. To further improve detection performance, a practical solution is to enhance the algorithm by further incorporating CAN messages into the detection algorithm. Sample CAN data, including bus speed, deceleration (can be calculated from speed), and brake switch associated with two sample events on May 7, 2021, are shown in Figure 5. Throttle percentage data were collected as well, but not shown in the figure because they were zero in both events.

Table I Near-crash detection evaluation results

| # of videos | # of events | TP | FP | FN | F1 Score | FPS |
|---|---|---|---|---|---|---|
| 5000 | 500 | 496 | 8 | 4 | 0.988 | 18 |

### E. Practical Issues and Event Location Mapping

While in the local test, Jetson processed the local videos frame by frame; in the real-world test, different camera hardware, settings, and different software design resulted in different frame-reading speed and stability. This was why the video reading function was designed as an individual thread. Also, when doing the bounding box size regressions, the system included the corresponding time for each value (height, width, and center) because the intervals between each pair of neighboring frames may not be uniform. Moreover, the camera type may influence system performance. About 2s latency in the video feed on the bus was noticed due to the use of IP camera connected via ethernet cables to the edge computing system. Jetson TX2 does not support auto boot-up. An external circuit driven by Arduino board was developed to automatically boot up the system.

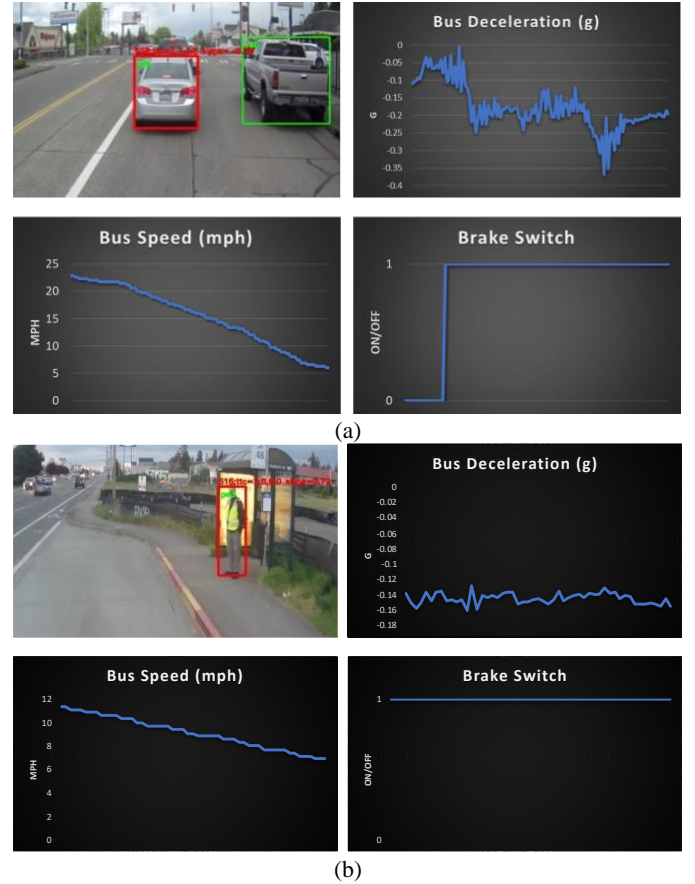

Fig. 5 Sample CAN data collected from (a) a vehicle-vehicle near-crash and (b) a vehicle-pedestrian near-crash on May 7[th], 2021.

It was also observed that the GPS coordinates collected were not in any of the standard formats. It took some time to figure out there was a linear relationship between the raw GPS coordinate and the WGS84 coordinate format. The conversion is shown as follows in Eq. (9) and we hope this information will be helpful to others planning to use the same GPS receiver.

$$\begin{cases} Lat_{WGS84} = 1.666 \times Lat_{raw} - 31.30174 \\ Lon_{WGS84} = 1.666 \times Lon_{raw} + 81.25186 \end{cases} \quad (9)$$

Figures 5, 6, and 7 present sample image, CAN, and GPS data collected in the real-world experiment by buses and cars. Figure 6 included three events near the University of Washington (UW) campus. The first one was on campus with a car, and the second one was on the 15th Street in the University District with a King County Metro bus, and the last one was west of campus near University Village. The GPS trajectory and event location data are valuable sources for analyses such as hotspot mapping and clustering. Figure 7 showed the trajectories and two near-crash events' spots on the OpenStreet Map with the corrected GPS coordinates during a trip on the UW campus. Figure 8 displayed all the near-crash events in October for Pierce Transit buses #230, #232, and #233, classified by bus identity and near-crash type (vehicle-vehicle or vehicle-pedestrian).

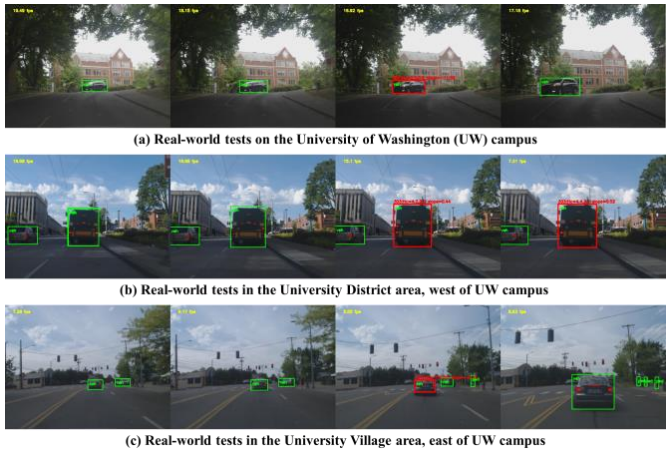

Fig. 6 Three near-crashes captured around UW area.

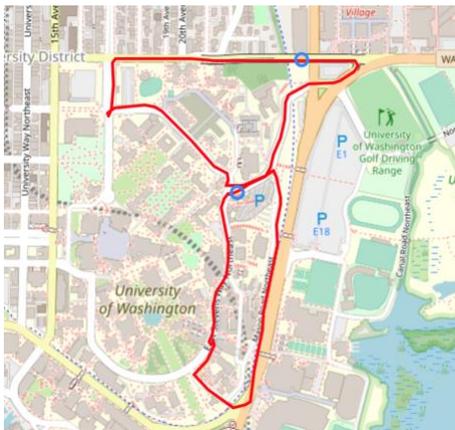

Fig. 7 The mapping of sample GPS trajectories (red curves) and near-crashes (blue circles) collected by cars.

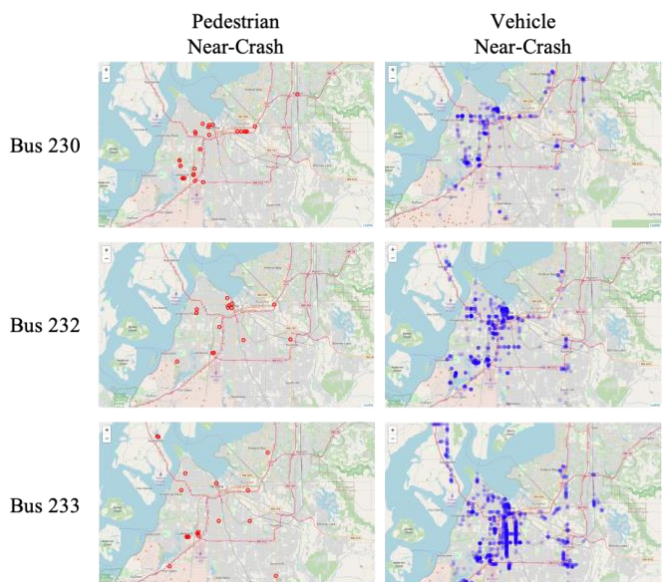

Fig. 8 The locations of pedestrian-related near-crashes (red circles) and vehicle-related near-crashes (blue circles) in October 2020.

*F. Comparison and Discussion*

This sub-section compares the proposed study qualitatively to the state-of-the-arts on using front-facing cameras for near-crash detection. The comparison is presented in Table II. The state of the arts has fully automated the process of near-crash detection and data collection using regular computers, but this study is among the first efforts to adopt edge computing, design and implement a system running on edge devices (Nvidia Jetson TX2). Regarding near-crash detection methods, the state of the arts tends to use machine learning, especially deep learning models. Convolutional neural network (CNN), long short-term memory (LSTM) neural network, and attention mechanism appear to be a good combination demonstrating superiority some most recent studies [28], [29]. Ibrahim et al. also showed that a bi-directional LSTM with self-attention mechanism performed better than single LSTM with regular attention [29]. However, these existing methods are more of black-box models due to the stack with multiple complicated deep learning modules thereby leading to limited efficiency, scalability, transferability, and interpretability.

It is okay in many projects just to leave the program running on regular computers and wait the near-crash extraction to be done, but large-scale near-crash detection do require real-time processing to filter out irrelevant videos and other data as soon as possible to significantly save transmission bandwidth, disk storage, and post-processing time. The proposed system is among the first to achieve real-time near-crash detection with the designed algorithms, system architecture, and the concept of edge computing. The proposed method is not sensitive to camera parameters or labeled near-crash data, thus it has great transferability to different dashcams and a good chance to detect the types of corner cases not covered by the training dataset, which is often limited to small scale in time and space.

The state of the arts was thoroughly validated with sufficient data, some has used thousands of video clips for validation purposes. In [30] and [28], the researchers used not only videos but also the telematics data such as acceleration and vehicle speed as part of the input, which indicated improved detection accuracy. In this study, we used online videos collected from different websites and unknown cameras for testing and finetuning the system and selected 5,000 video clips for validation. Then we deployed six of the devices on two cars and four buses since Summer 2020. The four devices on the four buses are still running upon the time we prepared this manuscript (June 2021). We have collected several hundred Gigabytes vehicle-vehicle and vehicle-pedestrian near-crash videos and data, which were all filtered and transmitted to the cloud server in real-time. These data are also expected to be valuable for multiple research topics such as traffic safety analysis and autonomous vehicle's corner case study.

The output data in most studies are videos, road user types, and risk levels associated with the events. Taccari et al. [30] estimated the TTC using a similar method with ours, but their estimation were based on solely two frames and without a differentiation between using height or width of the bounding box since that was not their focus. The output of our method firstly includes TTC, road user type, horizontal motion, and because of the real-time processing and CAN communication, it also collects the timestamp, latitude and longitude, speed, deceleration, brake switch, and the throttle data. The accuracy is not directly comparable among the studies given the lack of a widely accepted benchmark dataset and the difference in processing unit, input and output data, and model specifications, but we list them in the table for reference.

In summary, this study's prospective of application is very encouraging. It innovates in data collection by enabling real-time video analytics on the network edge and being backward compatible with existing vehicles. It addresses a few major concerns that are tied to some of the most critical research topics in smart transportation, such as the lack of vulnerable road user safety data, the bottleneck of going large scale in corner case collection for AV testing, and bridging the gap between theory/simulation and practice in transportation.

Table II Comparison with the state of the arts

| *Research work* | Ke et al. 2017 [7] | Kataoka et al., 2018 [6] | Taccari et al. 2018 [30] | Yamamoto et al. 2020 [28] | Ibrahim et al. 2021 [29] | **This study** |
|---|---|---|---|---|---|---|
| *Processing unit* | Regular computer | Regular computer with GPU | Regular computer with GPU | Regular computer with GPU | Regular computer with GPU | **Nvidia Jetson TX2** |
| *Edge computing* | No | No | No | No | No | **Yes** |
| *Key methods* | HOG, SVM, Optical Flow | Two-stream CNN, Semantic Flow | YOLO3, Optical Flow, Random Forest | CNN, LSTM, Attention | CNN, Bi-LSTM, Self-Attention | **Modeling deep-learning-generated object bounding boxes, SSD, SORT** |
| *Real-time processing* | No | No | No | No | No | **Yes** |
| *Need camera calibration or labeled near-crash data* | Yes | Yes | Yes | Yes | Yes | **No** |
| *Experimental data* | 30 hours of video | 6,200 video clips | SHRP 2 data with videos and telematics data | 4,200 video clips (15s each) and telematics data | 74,477 sequential frames | **Online videos, two cars, four buses; 5,000 video clips for validation and over 10,000 hours of real-world testing** |
| *Output near-crash data* | Pedestrian-related near-crashes | Risk level (high or low), road user type | Risk level (crash, near-crash, safe event), TTC, road user type | Near crash type in five risk levels, road user type | Near-crash label | **TTC, road user type, horizontal motion, timestamp, event location, vehicle trajectory, speed, deceleration, brake switch, throttle** |
| *Accuracy* | 0.900 | 0.645 | 0.870 | Confusion matrix | 0.994 | **0.988** |

## VI. CONCLUSION

In this paper, we introduced the motivation, design, development, and evaluation of an edge computing system for real-time near-crash detection and data collection. The proposed system was driven by real-time video analytics on IoT devices using existing dashcams. With the designs system-wise and algorithm-wise, this study addressed several key challenges in traffic near-crash detection and was among the first efforts in integrating edge computing with traffic video analytics and near-crash detection. Thorough experiments and analyses were conducted with recorded videos and real-world testing on cars and buses. The results were promising, demonstrating the potential of the proposed system for large-scale deployment with advantages including low cost, real-time processing, high accuracy, and great compatibility to different vehicles and cameras. The system can filter out irrelevant events in real-time, largely save bandwidth, computing, and storage resources, and increase the near-crash event extraction efficiency. It also increases the output data diversity; the data is expected to be very valuable sources for smart transportation applications such as by serving as surrogate data for traffic safety studies and as corner case data for automated vehicle testing research.

## ACKNOWLEDGMENT

The authors would like to thank the Federal Transit Administration (FTA) and the Pacific Northwest Transportation Consortium (PacTrans) for funding this research. They express gratitude to their research partners (Pierce Transit, WSTIP, DCS Technology Inc., VTTI, Volpe Lab, CUTR, Veritas, Dr. Jerome Lutin, Ms. Janet Gates, etc.) in the FTA project team for their invaluable contributions. They also thank the editors and reviewers for volunteering their time to review this paper.

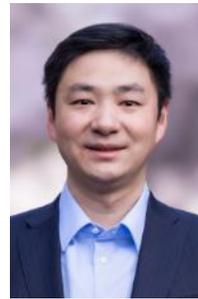

**Ruimin Ke** (Member, IEEE) is an Assistant Professor of civil engineering (smart cities) at the University of Texas at El Paso. He received his Ph.D. and master's degrees in civil engineering (transportation) at the University of Washington in 2020 and 2016, respectively. He received his B.E. degree in automation from Tsinghua University in 2014. Dr. Ke's research interest lies in intelligent transportation systems and smart cities with focuses on video image processing, machine learning, and the Internet of Things applications.

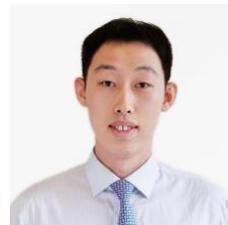

**Zhiyong Cui** (Member, IEEE) is a Postdoctoral Research Associate working at the Smart Transportation Applications and Research (STAR) Lab at the University of Washington (UW). He's also a UW Data Science Postdoctoral Fellow at the eScience Institute. He received the B.S. degree in software engineering from Beihang University in 2012, the M.S. degree in software engineering and microelectronics from Peking University in 2015, and his Ph.D. in civil engineering in 2021. Dr. Cui's primary research focuses on deep learning, machine learning, urban computing, traffic forecasting, connected vehicles, and transportation data science.

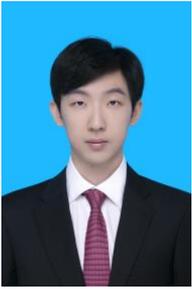
**Yanlong Chen** received the B.E. degree in mechanical engineering from Tsinghua University, Beijing, China, in 2019. He is currently pursuing the M.E. degree in mechanical engineering at the University of Tokyo. In 2019, he was a visiting student in the Smart Transportation Application and Research Laboratory (STAR Lab), University of Washington. His research interests include computer vision and its application in robot manipulation.

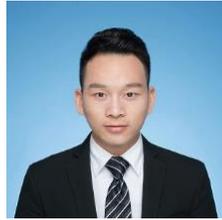
**Meixin Zhu** is a Ph.D. student at the University of Washington. He also serves as a research assistant in Smart Transportation Applications and Research Laboratory (STAR Lab) at the University of Washington. He received the BSc and MSc degrees in Traffic Engineering in 2015 and 2018 respectively from Tongji University. Zhu's research interests include autonomous driving, artificial intelligence, big data analytics, driving behavior, traffic-flow modeling and simulation, and naturalistic driving study.

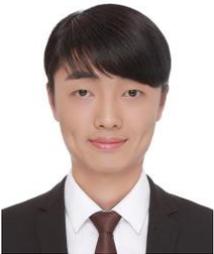
**Hao (Frank) Yang** (Student member, IEEE) received the B.S. degree of Telecommunication Engineering from both Beijing University of Posts and Telecommunications (2017) and University of London (2017). He is currently a Ph.D. student at the Smart Transportation Research and Application Lab (STAR Lab), Department of Civil and Environmental Engineering, University of Washington. He is an associate editor for IEEE ITSC, reviewer of Conference on Computer Vision and Pattern Recognition (CVPR) 2019, 2020, Transportation Research Part C: Emerging Technologies and IEEE Transactions on Intelligent Transportation Systems and etc. His research interests are focused on transportation data analysis, deep learning and computer vision.

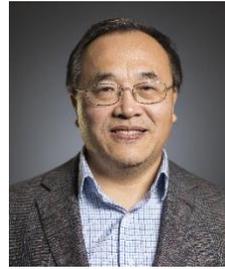
**Yinhai Wang** (Senior Member, IEEE) received the master's degree in computer science from the University of Washington (UW) and the Ph.D. degree in transportation engineering from the University of Tokyo, in 1998. He is currently a Professor in transportation engineering and the Founding Director of the Smart Transportation Applications and Research Laboratory (STAR Lab), UW. He serves as the Director of the Pacific Northwest Transportation Consortium (PacTrans), USDOT University Transportation Center for Federal Region 10. Dr. Wang is the Chair of the Artificial Intelligence and Advanced Computing Committee of the Transportation Research Board. He is a fellow with American Society of Civil Engineers (ASCE) and the Past President of the ASCE Transportation and Development Institute (T&DI). He is also a member of the IEEE Smart Cities Technical Activities Committee and was an elected member of the Board of Governors for the IEEE ITS Society from 2010 to 2013.